\documentclass[10pt,twocolumn,letterpaper]{article}

\usepackage[pagenumbers]{cvpr} %

\usepackage[dvipsnames]{xcolor}

\usepackage{tabularx}
\usepackage{xcolor}

\usepackage{verbatim}
\usepackage{comment}

 \usepackage{array,multirow,graphicx}
 \usepackage{float}
 \usepackage[accsupp]{axessibility} %

\newcommand{\zgeo}{\mathbf{z}_{\texttt{geo}}}
\newcommand{\zapp}{\mathbf{z}_{\texttt{app}}}
\newcommand{\zexp}{\mathbf{z}_{\texttt{exp}}}

\newcommand{\Fgeo}{\mathcal{F}_{\texttt{geo}}}
\newcommand{\Fapp}{\mathcal{F}_{\texttt{app}}}
\newcommand{\Fexp}{\mathcal{D}_{\texttt{exp}}}

\definecolor{cvprblue}{rgb}{0.21,0.49,0.74}
\usepackage[pagebackref,breaklinks,colorlinks,citecolor=cvprblue]{hyperref}

\def\paperID{16417} %
\def\confName{CVPR}
\def\confYear{2024}

\newcommand{\OURS}{\texttt{Mono}NPHM}

\title{\OURS: Dynamic Head Reconstruction from Monocular Videos}

\author{
Simon Giebenhain$^1$ \quad
Tobias Kirschstein$^1$ \quad
Markos Georgopoulos$^{2*}$ \quad 
Martin Rünz$^2$ \\
Lourdes Agapito$^3$ \quad
Matthias Nie{\ss}ner$^1$ \vspace{0.2cm}\\
$^1$Technical University of Munich  \qquad $^2$Synthesia \qquad $^3$University College London
}

\begin{document}

\twocolumn[{
\renewcommand\twocolumn[1][]{#1}
\maketitle
\thispagestyle{empty}
\begin{center}
  \newcommand{\teaserwidth}{\textwidth}
   \vspace{-0.4cm}

  \centerline{
    \includegraphics[width=\teaserwidth]{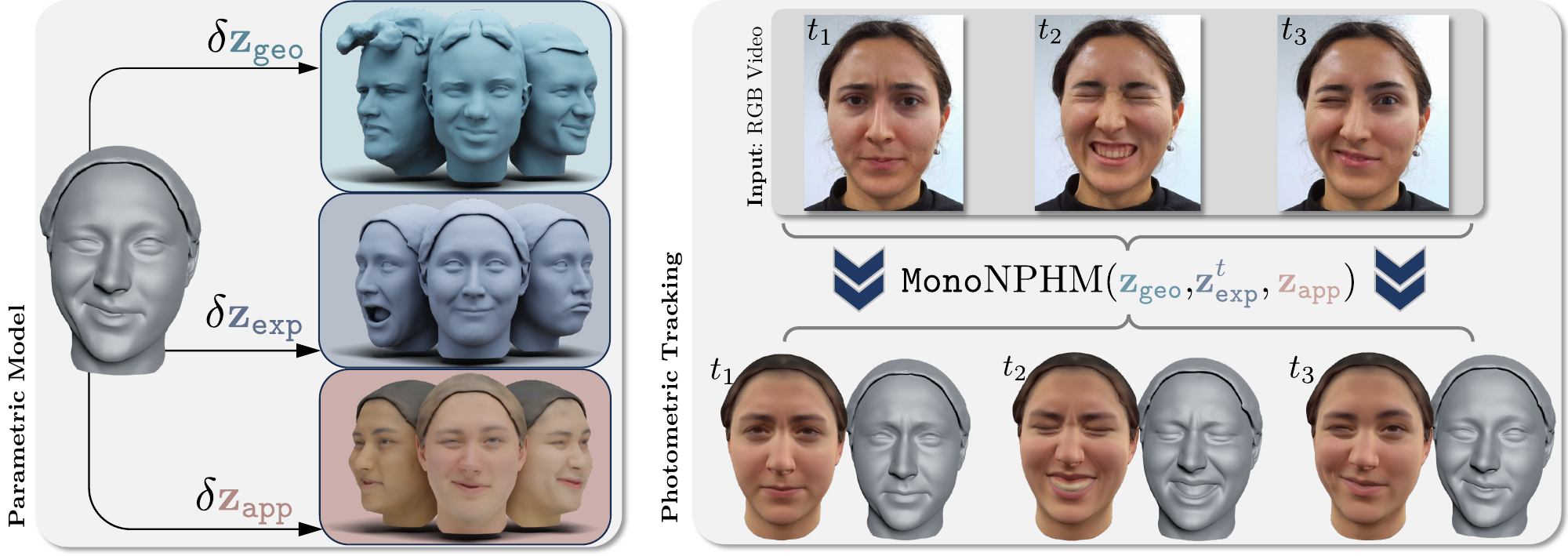}
    }
     \vspace{-0.2cm}
    \captionof{figure}{
    We present \OURS, a neural-field-based parametric head model (left), for dynamic 3D head reconstruction in monocular videos (right).
    On the left we demonstrate the effect on a reconstructed human head, by individually varying shape (top box), expression (middle box) and appearance (bottom box).
    The right-hand side illustrates three input RGB frames (top row), and our reconstructed geometry (bottom row). We also show the reconstructed appearance under estimated lighting conditions, which is the basis of our reconstruction.
    }
  \label{fig:teaser}
      \vspace{-0.2cm}

 \end{center}
}]

\begin{abstract}

We present Monocular Neural Parametric Head Models (\OURS) for dynamic 3D head reconstructions from monocular RGB videos. 
To this end, we propose a latent appearance space that parameterizes a texture field on top of a neural parametric model. %
We constrain predicted color values to be correlated with the underlying geometry such that gradients from RGB effectively influence latent geometry codes during inverse rendering.
To increase the representational capacity of our expression space, we augment our backward deformation field with hyper-dimensions, thus improving color and geometry representation in topologically challenging expressions.
Using \OURS~ as a learned prior, we approach the task of 3D head reconstruction using signed distance field based volumetric rendering. 
By numerically inverting our backward deformation field, we incorporated a landmark loss using facial anchor points that are closely tied to our canonical geometry representation.
To evaluate the task of dynamic face reconstruction from monocular RGB videos we record 20 challenging Kinect sequences under casual conditions.
\OURS~ outperforms all baselines with a significant margin, and makes an important step towards easily accessible neural parametric face models through RGB tracking.

{\let\thefootnote\relax\footnotetext{\scriptsize{Website: \url{https://simongiebenhain.github.io/MonoNPHM}}

* Work done while MG was at Synthesia.}}
\end{abstract}

\section{Introduction}
\label{sec:intro}

Tracking, animation, and reconstruction of human faces and heads under complex facial movements are fundamental problems in many applications such as computer games, movie production, telecommunication, and AR/VR settings. 
In particular, obtaining high-fidelity 3D head reconstructions from monocular input videos is a common scenario in many practical settings, e.g., when only a commodity webcam is available.

Recovering the 3D head geometry throughout a monocular RGB video, however, is inherently under-constrained. The task is further complicated in the presence of depth ambiguity, complex facial movements, and strong lighting and shadow effects. 
Therefore, to disambiguate the 3D scene dynamics, it is common to introduce a set of assumptions about plausible facial structure, expressions, and appearance, often in the form of a model prior.

To regularize this otherwise heavily under-constrained problem, 
the most widely adopted model-prior, are 3D morphable models (3DMMs) \cite{blanz1999morphable}, which capture shape, expression, and appearance variations through the use of principal component analysis (PCA) over a dataset of 3D scans that have been registered with a template mesh. 
Therefore, their expressiveness is often limited by the underlying (multi-)linear statistical model, the resolution of the template mesh, and its topology. 
Recent neural variants of mesh-based 3DMMs~\cite{COMA:ECCV18,SpiralNet++,FaceScape, FaceVerse,Lattas_2023_CVPR_fitme} and neural-field-based parametric face models~\cite{i3DMM, ImFace, giebenhain2023nphm,phomoh,lin2023ssif} constitute more detailed model priors, but so far do not tackle 3D head reconstruction from monocular RGB videos.

In this work, we propose \OURS, a neural parametric head model tailored towards monocular 3D reconstruction from RGB videos.  
We model an appearance field, coupled with a signed distance field (SDF) that represents the geometry, in \emph{canonical space}. Facial expressions are represented using a backward deformation field that establishes correspondences from \emph{posed space} into the canonical space.
Additionally, we augment our backward deformation model using hyper-dimensions~\cite{hypernerf}, in order to increase the dynamic capacity of our model. 
Building on top of our parametric model, we perform photometric 3D head tracking, by optimizing for latent geometry, appearance, and expression codes.
To establish an RGB loss, we utilize SDF-based volumetric rendering \cite{wang2021neus} of rays in posed space which are backward-warped into canonical space. To account for different lighting conditions we incorporate spherical harmonics shading \cite{spherical_harmonics} into the volumetric rendering.
Additionally, we find that a landmark loss is crucial for robust tracking through extreme facial movements.
We use a discrete set of facial anchor points that is tightly coupled with our geometry representation \cite{giebenhain2023nphm}. 
We forward-warp the anchors by numerically inverting our backward deformation field using iterative root finding \cite{chen2021snarf} and project them into image space to compute our landmark loss. 

Compared to our strongest baselines we improve the reconstruction fidelity, measured by Chamfer distance, by 20\%. To sum up our contributions are as follows:
\begin{itemize}
    \item We introduce \OURS, a neural parametric head model that jointly models appearance, geometry, and expression and is augmented with hyper-dimensions for an increased dynamics capacity.
    \item We tightly condition our appearance network on the underlying geometry, to allow for meaningful gradients during inverse rendering, which we formulate based on dynamic volume rendering of implicit surfaces. 
    \item We introduce a landmark loss using discrete facial anchor points that are tightly coupled with our implicit geometry.

\end{itemize}

\newcolumntype{Y}{>{\centering\arraybackslash}X}
\newcolumntype{P}[1]{>{\centering\arraybackslash}p{#1}}
\begin{figure*}[htb]
    \vspace{-0.2cm}

    \centering
    \includegraphics[width=\textwidth]{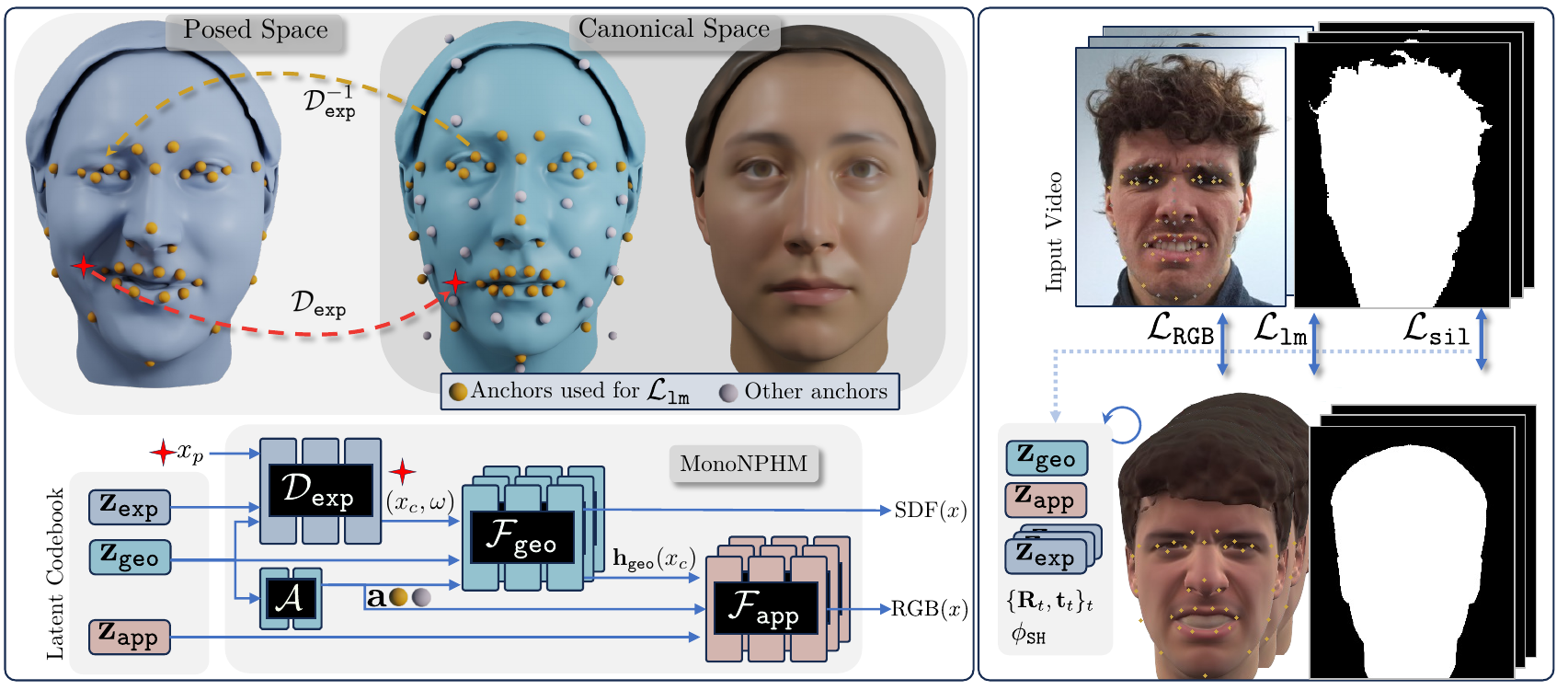}
     \begin{tabularx}{\textwidth}{
         P{0.62\textwidth}
         P{0.32\textwidth}
     }
    a) \OURS   & b) Tracking\\
    \end{tabularx}
    \vspace{-0.6cm}
    \caption
    {\textbf{Method overview}: 
    (a) Shows how \OURS~operates: First, points $x_p$ in posed-space are backward-warped through $\Fexp$ into canonical space (indicated by red star and arrow). 
    Our canonical geometry and appearance fields are conditioned on facial anchors $\mathbf{a}_c$ (yellow and gray points in canonical space).
    Instead of conditioning $\Fapp$ on canonical coordinates $(x_c, \omega)$, we use hidden features $\mathbf{h}_{\texttt{geo}}$ extracted from the geometry network.
    (b) We approach tracking using SDF-based volumetric rendering \cite{wang2021neus} to build photometric and silhoutte terms. Additionally, we enforce a landmark loss by numerically inverting $\Fexp$ using iterative root finding (as indicated by the yellow arrow on the left).
    }
    \label{fig:method}
        \vspace{-0.2cm}

\end{figure*}

\section{Related Work}
\label{sec:related_work}

\paragraph{Mesh-based face models}
Starting with the seminal work on 3DMMs~\cite{blanz1999morphable}, template-mesh-based PCA models \cite{paysan20093d, FLAME, hifi3dface2021tencentailab,HACK,booth2018large} have been widely used for many application in computer graphics and vision. %
To relax the rigid linear assumption of PCA, subsequent efforts utilized variation auto-encoders(VAEs)~\cite{Kingma2014_vae}, generative adversarial networks (GANs)~\cite{goodfellow2014GANs}, and diffusion models~\cite{ho2020denoising} to replace the PCA-basis underlying classical mesh-based 3DMMs \cite{COMA:ECCV18,SpiralNet++,FaceScape,FaceVerse,Lattas_2023_CVPR_fitme,galanakis2023fitdiff,Paraperas_2023_ICCV_relightify}.

\paragraph{3D face reconstruction from RGB}
Reconstructing the 3D geometry of a head from RGB images or videos is a fundamental problem in computer vision. Standard approaches optimize the parameters of a 3DMM based on the 2D input \cite{bai2023ffhq,gecer2019ganfit,FaceScape,thies2016face}.
Optimizing the parameters from arbitrary poses, especially in the presence of occlusions and strong shadows, is a very challenging problem. Learning-based methods address this issue by training neural networks to predict the face representation from the input images \cite{danvevcek2022emoca,sanyal2019learning,tewari2017mofa,lin2020towards,DECA}.
In order to model details beyond the 3DMM template, such as wrinkles, several efforts utilize shape-from-shading \cite{garrido2016reconstruction,jiang20183d,suwajanakorn2014total} while others model facial details as displacements maps \cite{chen2019photo,feng2021learning,lei2023hierarchical,10.1145/964965.808594}.
Instead of relying on a fixed-topology template mesh, we approach 3D reconstruction from RGB inputs using neural-field-based parametric head models, allowing for the representation of complete human heads with varying topologies.

\paragraph{Neural field-based face models}
Recent advances on neural fields \cite{survey_neural_fields}, have shown impressive results on geometry reconstruction and generation \cite{park2019deepsdf, mescheder2019occupancy, peng2020convolutional, yariv2021idr,wang2021neus}, neural radiance fields (NeRFs) \cite{mildenhall2021nerf, mueller2022instant, Chen2022ECCV_tensorf,kerbl3Dgaussians}, and dynamic scene reconstructions \cite{park2021nerfies, hypernerf,kirschstein2023nersemble,nerfplayer, hyperreel,icsik2023humanrf}.
Such techniques have been recently used in the context of 3D generative models \cite{pi-GAN, Chan2022_eg3d, Wang_2023_CVPR_rodin, bergman2022gnarf}, and NeRF-based parametric models \cite{wang_2022_morf, hong2021headnerf,zhuang2022mofanerf, Buhler_2023_ICCV_preface, auth_ava} to generate high-fidelity heads that can be rendered from different views.
Others have focused on highly detailed geometry representation \cite{i3DMM, ImFace, giebenhain2023nphm, phomoh, zheng2023imface++}, design a diffusion prior for robust reconstruction from depth sensors \cite{tang2023dphms}, and facilitate few-shot 3D reconstruction from RGB images using a mixture of model-based fitting and test-time fine-tuning of model parameters \cite{ramon2021h3d, lin2023ssif, Buhler_2023_ICCV_preface}.
Closer to our work, \cite{lin2023ssif} is able to reconstruct an animatable head avatar from a single image in the wild.
In this work, however, we focus on dynamic 3D reconstruction from monocular RGB videos by explicitly modeling the deformations, which allows us to obtain correspondences across the video.

\paragraph{Person-specific head avatars}
To escape the limitation of generalizing parameter space, methods for \emph{person-specific} avatars from monocular videos, have shown impressive results. These methods usually incorporate a 3DMM to introduce control in the neural implicit representation \cite{gafni2021nerface, Athar_2022_CVPR_rignerf, zielonka2022instant, IMavatars,
Zheng2023pointavatar,
lombardi2021mixture,qian2023gaussianavatars}.
However, they lack generalization and as such, require to be trained per video, in contrast, to our method that generalizes across identities and expressions.

A different line of work focuses on the construction of photo-realistic avatars from abundant multi-view video recordings \cite{lombardi2021mixture,qian2023gaussianavatars,kirschstein2023diffusionavatars,saito2023relightable}.

\section{\OURS}
\label{sec:method}

Our work aims at dynamic 3D Face reconstruction in monocular RGB videos. We approach this heavily under-constrained task using model-based photometric tracking through inverse SDF-based rendering. 
In this section we describe the construction of our underlying, neural field-based model, \OURS, illustrated in \cref{fig:method}, along with its disentangled parametric spaces for shape (\cref{sec:canonical_geo}), appearance (\cref{sec:canonical_app}) and expression information (\cref{sec:dynamics}).
In \cref{sec:tracking} we propose a model-based dynamic 3D reconstruction algorithm, based on \OURS.

\subsection{Canonical Geometry Representation}
\label{sec:canonical_geo}

We represent the head geometry in canonical facial expression, as described by latent code $\zgeo$, using a neural SDF 
\begin{equation}
\mathcal{F}_{\texttt{geo}}: \mathbb{R}^{3 + d_{\texttt{geo}}} \rightarrow \mathbb{R}^1, x_c \mapsto \text{SDF}(x_c),
\end{equation}
operating on points $x_c$ in canonical space. Such an implicit representation provides the necessary topological flexibility to describe complete heads, including hair.

We follow NPHM~\cite{giebenhain2023nphm} and compose $\Fgeo$ as an ensemble of local MLPs
\begin{equation}
    \mathcal{F}_{\texttt{geo}}(x_c, \mathbf{z}_{\texttt{geo}})\!=\! \sum_{k\in\mathcal{N}_{x_c}} w_k(x_c, \mathbf{a}_c^k) f_{\texttt{geo}}^k(x_c\!-\!\mathbf{a}^k_c; \mathbf{z}_{\texttt{geo}}),
    \label{eq:local_blending}
\end{equation}
which are centered around facial \emph{anchor} points $\mathbf{a}_c^k = \mathcal{A}(\zgeo) \in \mathbb{R}^{65\times3},$ that are predicted by a small MLP $\mathcal{A}$ 
based on the geometry code $\mathbf{z}_{\texttt{geo}}$. Therefore, the anchor positions constitute an integral part of the pipeline and provide an important discrete structure which we leverage as a landmark loss for monocular tracking in \cref{sec:landmark_loss}. 

For this purpose we design an anchor layout consisting of $65$ points, s.t. the most important landmarks of common detectors coincide with anchor points, as shown in \cref{fig:method}. 
To account for the increased number of anchors, we restrict the computation to the bipartite $k$NN-graph from $x_c$ to its $8$ nearest anchors $\mathcal{N}_{x_c}$, Compared to NPHM, which evaluates all local MLPs, in this case 65, this is more than an eight-fold reduction in memory.
To account for the non-uniform spatial arrangement of anchors, we re-scale $w_k$ for each neighborhood separately. Details are provided in our supplementary material.

\subsection{Canonical Appearance Representation}
\label{sec:canonical_app}

We model appearance changes between subjects using separate latent codes $\zapp$, that condition a texture field $\Fapp$.

To emphasize the dependence of appearance on the geometry, we incorporate a strong connection between the two networks, similar to PhoMoH~\cite{phomoh}. 
Our motivations come from the fact that an appearance space, which is completely independent of the geometry, could reconstruct the observed color images without providing meaningful gradients for the latest geometry codes.

To this end, we condition our texture field $\Fapp$, on features $\mathbf{h}_{\texttt{geo}}(x_c) \in \mathbb{R}^{16}$ which are extracted from the last layer of the geometry MLP $\Fgeo$ using two narrow linear layers.
As illustrated in \cref{fig:method}, $\Fapp$ follows the same local structure as our geometry network, i.e. local appearance MLPs
\begin{equation}
    f_{\texttt{app}}^k(\mathbf{h}_{\texttt{geo}}^k(x_c); \mathbf{z}_{\texttt{app}}) \in [0,255]^3
    \label{eq:color_communication}
\end{equation}
are blended using the same weights as in \cref{eq:local_blending}.
As we will show later in \cref{sec:ablations}, removing the dependence of $\Fapp$ on spatial coordinates $x_c$ and using features $\mathbf{h}_{\texttt{geo}}(x_c)$ instead, is beneficial for RGB-based 3D reconstruction.

\subsection{Representing Dynamics}
\label{sec:dynamics}

While both previous components operate in canonical space, it is the task of our deformation network 
\begin{equation}
    \Fexp: \mathbb{R}^{3+d_{\texttt{exp}}+d_{\texttt{geo}}} \rightarrow \mathbb{R}^3, x_p \mapsto x_c
\end{equation}
to \emph{backward}-warp points $x_p$ in posed space into canonical coordinates $x_c$. 
Such a formulation implies that all changes in the geometry and appearance fields between two expressions can be explained through a deformation of space.

To relieve this strong assumption, we relax the formulation by adding \emph{hyper-dimensions}, or \emph{ambient dimensions},~\cite{hypernerf} to the output of the deformation network, i.e. $\Fexp(x_p; \mathbf{z}_{\texttt{exp}}, \mathbf{z}_{\texttt{geo}}) = (x_c, \omega) \in \mathbb{R}^{3+h}$, where $h$ is the number of hyper-dimensions (in practice we use $h=2$). Consequently, $\Fgeo$ is provided with canonical coordinates \emph{and} hyper-coordinates $\omega$, which increase the dynamic capacity of the overall network. 
\cref{fig:effect_hyper} demonstrates the topological issues that arise without using hyper-dimensions.

Following previous work \cite{NPM, giebenhain2023nphm}, $\Fexp$ is conditioned on both $\zexp$ \emph{and} $\zgeo$ since the identity information is relevant to find correct correspondences between $x_p$ and $x_c$. Note that NPHM~\cite{giebenhain2023nphm} uses forward deformations, which we ablate to perform inferior compared to formulation.

\subsection{Training}
\label{sec:training}

We train all model components and latent codes end-to-end using an auto-decoder formulation~\cite{park2019deepsdf}. 
Given a public dataset consisting of high-quality textured 3D scans~\cite{giebenhain2023nphm}\footnote{We use the version 2 release containing a total of 473 identities}, 
we sample points $x_p$ near the mesh surface and pre-compute $\text{SDF}(x_p)$ and $\text{RGB}(x_p)$ values for direct supervision of our geometry and color fields. Conceptually, we optimize for model parameters $\Theta$ and latent codes $\mathcal{Z}$
\begin{eqnarray}
        \underset{\mathcal{Z}, \Theta}{\mathrm{argmin}}\!&\sum\limits_{s\in S, e \in E_s} \lambda_{\text{SDF}}| \Fgeo(\Fexp(x_p)) - \text{SDF}(x_p) | +  \nonumber \\
        &\lambda_{\text{RGB}}| \Fapp(\mathbf{h}_{\texttt{geo}}(x_c)) - \text{RGB}(x_p) | + \lambda_{\text{reg}}\mathcal{L}_{\text{reg}},\quad
\end{eqnarray}
where $\mathcal{L}_{\textbf{reg}}$, among others, imposes regularization on all latent codes, supervises anchor predictions, and regularizes deformations and predicted hyper-dimensions to be small. We provide more details about the training and network architectures in our supplementary document.

\section{3D Dynamic Face Reconstruction}
\label{sec:tracking}

Our main goal is tracking heads in the parametric space of \OURS, in the case of a single, monocular RGB input video.
In such a challenging scenario, it is essential that a strong, but expressive, model-prior can guide the optimization through the often under-constrained task. We conceptually visualize this task in \cref{fig:method}.
Given a video sequence of RGB frames $\{ I_1, \dots I_T\}_{t=1}^T$, associated silhouettes $\{ S_1, \dots S_T\}_{t=1}^T$ and 2D facial landmarks $\{L_1, \dots, L_T\}_{t=1}^T$ we aim to reconstruct model parameters $\phi = \{\zapp, \zgeo\} \cup \{\zexp\}_{t=1}^T$, composed of time-invariant codes $\zgeo$ and $\zapp$, as well as, per frame expression codes $\zexp^{t}$. We solve the tracking task by minimizing the energy
\begin{equation}
    \underset{\phi, \zeta, \Pi}{\mathrm{argmin}} 
    \sum_{t=1}^T
    \mathcal{L}_{\texttt{RGB}}^t + \lambda_{\texttt{sil}}\mathcal{L}_{\texttt{sil}}^t + 
    \lambda_{\texttt{lm}}\mathcal{L}_{\texttt{lm}}^t  + 
    \lambda_{\texttt{reg}}\mathcal{L}_{\texttt{reg}}^t
    \label{eq:tracking}
\end{equation}
with respect to latent codes $\phi$, head poses $\Pi=\{\mathbf{R}_t, \mathbf{t}_t\}_{t=1}^T$, as well as, lighting parameters $\zeta \in \mathbb{R}^9$ of a 3-band spherical harmonics approximation~\cite{spherical_harmonics}. 

The data term of our energy contains a pixel-level RGB loss $\mathcal{L}_{\texttt{RGB}}$ and silhouette loss $\mathcal{L}_{S}$, as explain in \cref{sec:rendering_formulation,sec:shading,sec:rendering_losses}, and a landmark loss $\mathcal{L}_{\texttt{lm}}$ for coarse guidance of the expression (see \cref{sec:landmark_loss}). 
We describe our regularization term and optimization strategy in \cref{sec:regularization_loss} and \cref{sec:optimization_strategy}, respectively.

\subsection{Rendering Formulation}
\label{sec:rendering_formulation}

To relate the 3D neural field, parameterized by the latent codes, with the 2D observations, we perform volumetric rendering in posed space. 
Given intrinsic camera parameters $K$, we transfer the head pose into an extrinsics matrix $E_t = [\mathbf{R}_t|\mathbf{t}_t]$, and shoot a ray $r_p(\tau) = o + \tau \cdot d$ into the scene at time $t$. Samples along the ray are warped into canonical space 
$
r_c(\tau) = \Fexp(r_p(\tau); \mathbf{z}_{\texttt{exp}}^t, \mathbf{z}_{\texttt{geo}}).
$
Consequently, we infer SDF values $\mathcal{F}_{\texttt{geo}}(r_c(\tau);\mathbf{z}_{\texttt{geo}})$ in caonical space. 
For volume rendering, we rely on the formulation of NeuS~\cite{wang2021neus} to transfer SDF values along a ray into rendering densities $\sigma(r_c(\tau))$. In total, predicted RGB values $c(\tau)$ along a ray are aggregated into pixel colors 
\begin{equation}
\hat{I}_t(r)\!=\!\int_{\tau_n}^{\tau_f}\!w(\tau) c(\tau)\!d\tau
\end{equation}
using volume rendering \cite{raytracing_volume_densities, mildenhall2021nerf}. Here, rendering weights $w(\tau)$ and the accumulated transmittance $T(\tau)$ are defined as follows:
\begin{equation}
    w(\tau) = T(\tau)\sigma(r_c(\tau)),~\text{and}~T(\tau) = e^{-\int_{\tau_n}^{\tau}\sigma(r_c(s))ds }.
\end{equation}

\subsection{Spherical Harmonics Shading}
\label{sec:shading}
To bridge the domain gap between our albedo appearance space, and in-the-wild lighting effects, we include a 3-bands spherical harmonics as a simple approximation for the scene lighting~\cite{spherical_harmonics}. Thus, we obtain shaded RGB predictions
\begin{equation} c(\tau) = \text{SH}_{\zeta}(n(\tau)) \mathcal{F}_\texttt{app}\!\left(\textbf{h}_{\texttt{geo}}(r_c(\tau)); \mathbf{z}_{\texttt{app}}\right)
\end{equation}
by multiplying predicted colors with the spherical harmonics term, parameterized by $\zeta\in\mathbb{R}^9$. For this, we use world space normals
$
n(\tau) = \mathbf{R}_t\nabla_{x_p}\Fgeo(x_c; \zgeo),
$
where the dependence on $x_p$ is included in the relation $x_c = \Fexp(x_p; \zexp, \zgeo)$. 
We show the importance of accounting for lighting effects in \cref{sec:ablations}.

\subsection{Rendering Losses}
\label{sec:rendering_losses}

The most important term in our inverse rendering is the color loss $\mathcal{L}_{\texttt{RGB}}^t = \operatorname{MAE}(\hat{I}_t, I_t)$, which measures the average L1-loss over all pixels in the foreground region, between predicted image colors $\hat{I}_t$ and observed images $I_t$.

Additionally, we supervise the silhouette $S_t$ using an average binary cross-entropy loss $\mathcal{L}_{\texttt{sil}}^t = \operatorname{BCE}(\hat{S}_t(r), S_t(r))$ over all pixels, with predicted forground
$
\hat{S}_t = \int_{\tau_n}^{\tau_f} w(\tau)d\tau.
$

\subsection{Landmark Loss}
\label{sec:landmark_loss}

Next to the above-mentioned rendering losses we observe that the optimization can get stuck in local minima for extreme mouth movements. We address this issue by incorporating a landmark loss, a common practice in face tracking.

For this purpose, we exploit the structure of the underlying NPHM model that is offered through its anchor points $\mathcal{A}(\zgeo) = \mathbf{a}_c$.
We determine the anchor positions $\mathbf{a}_p^t$ in posed space that satisfy
\begin{equation}
 0 = {\mathbf{a}_c^t} - \Fexp({\mathbf{a}_p^t}; \zexp, \zgeo)
\end{equation}
using iterative root finding~\cite{chen2021snarf}, i.e. the backward deformation field is inverted through a numerical procedure.
To coarsely guide $\mathbf{z}_{\texttt{exp}}$ during tracking we enforce 
\begin{equation}
    \mathcal{L}_{\text{lm}} = \operatorname{MSE}(\pi_{K, E_t}({\mathbf{a}_p^t}), L_t),
    \label{eq:landmark loss}
\end{equation}
which measures the screen-space distance between detected landmarks $L_t$ and projected posed anchors,
where $\pi_{K, E_t}$ denotes a perspective projection using camera intrinsics $K$ and extrinsics $E_t$.

\subsection{Regularization}
\label{sec:regularization_loss}

We encourage the latent codes to stay within a well-behaved parameter range, which is also enforced during training:
\begin{equation}
    \mathcal{L}_{\text{prior}} = \Vert \zgeo \Vert^2 + \lambda_{\texttt{app}}\Vert \zapp \Vert^2 + \frac{\lambda_{\texttt{exp}}}{T}\sum_t\Vert \zexp \Vert^2.
\end{equation}
Additionally, we use the symmetry loss from NPHM~\cite{giebenhain2023nphm} on the local latent codes contained in $\zgeo$ and $\zapp$, and enforce temporal smoothness on time-dependent parameters
\begin{equation}
    \mathcal{L}_{\text{smooth}} = \text{TV}(\zexp) + 
    \lambda_{\text{rot}}\text{TV}(\mathbf{R}_t) + 
    \lambda_{\text{trans}}\text{TV}(\mathbf{t}_t).
\end{equation}

\subsection{Optimization Strategy}
\label{sec:optimization_strategy}

We optimize \cref{eq:tracking} using stochastic gradient descent (SGD) and the Adam optimizer~\cite{adam}. 
We initialize all latent codes as zeros, $\zeta$ is initialized as uniform lighting from all directions, and head poses $\mathbf{R}_t, \mathbf{t}_t$ are initialized from a tracked FLAME model.

We start our optimization by separately optimizing the first frame, and then optimize for the remaining frames sequentially in a frame-by-frame fashion, where $\zgeo$, $\zapp$, and $\zeta$ remain frozen.
This strategy provides good estimates over all parameters and serves as initialization for our main stage, where we optimize over \emph{all} parameters jointly. For each optimization step a random timestep $t$ and random rays for the rendering losses are sampled. Our smoothness loss is computed between the neighboring frames $t-1$ and $t+1$.

\section{Results}
\label{sec:experiments}

\newcolumntype{Y}{>{\centering\arraybackslash}X}
\newcolumntype{P}[1]{>{\centering\arraybackslash}p{#1}}
\begin{figure*}[htb]
    \vspace{-0.2cm}

    \centering
    \includegraphics[width=\textwidth]{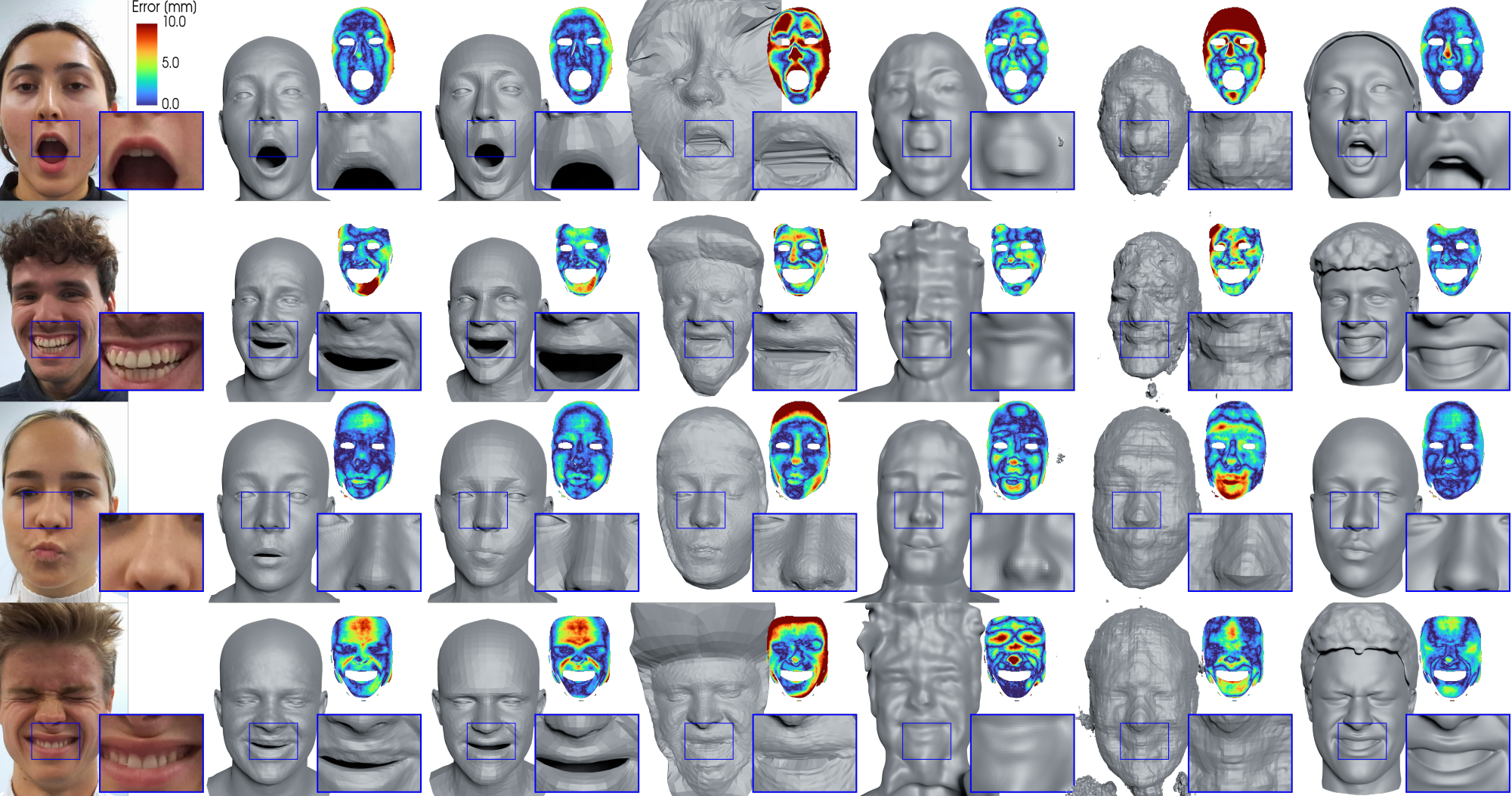}
     \begin{tabularx}{\textwidth}{
         P{0.11\textwidth}%
         P{0.11\textwidth}%
         P{0.12\textwidth}%
         P{0.14\textwidth}%
         P{0.12\textwidth}%
         P{0.12\textwidth}%
         P{0.10\textwidth}%
     }
    \small{Input RGB} & \small{DECA~\cite{DECA}} & \small{MICATracker~\cite{MICA_ECCV2022}} & \small{NHA~\cite{nha} }&  \small{IMavatar~\cite{IMavatars}} & \small{HeadNeRF~\cite{hong2021headnerf}}&\small{Ours}\\
    \end{tabularx}
        \vspace{-0.4cm}

    \caption{
    Comparison of the 3D reconstruction quality from monocular RGB videos against our baselines. 
    We show error maps with color-coded point-to-mesh distance from ground truth Kinect depth to the reconstructed meshes.
    }
    \label{fig:main_results}
        \vspace{-0.2cm}

\end{figure*}

To evaluate our goal of dynamic face reconstruction, we record 20 Kinect sequences in a casual setting, for a lack of publicly available alternatives. 
The RGB sensor serves as input, while the depth sensor allows for a geometric evaluation. 
We record 5 participants (3 female, 2 male) under a wide range of facial expressions, emotions and include one talking sequence. All participants signed agreements compliant with GDPR requirements. We record each sequence for 12 seconds at 15 frames per second, resulting in 180 frames per sequence.

\subsection{Metrics}

We report unidirectional $L_1$-Chamfer distance in meters from the back-projected depth map to the reconstructions, which cover the complete head.
Similarly, we report the unidirectional cosine-similarity of normals. %
Additionally, we measure the recall~\cite{what3d_cvpr19}, i.e. the percentage of ground truth points that are covered by at least one point on the reconstruction w.r.t. to a given threshold distance. 

\textbf{Evaluation Protocol ~}
To eliminate any remaining depth ambiguity, we optimize for a similarity transform from reconstructed mesh to ground truth point cloud using ICP~\cite{ICP}. 
To exclude the sensor noise of the regions inside the mouth and eyes and to account for differences between the compared methods, we remove these regions, as well as the hair and neck region, using facial segmentation~\cite{zheng2022farl}. 
We visualize the resulting ground truth point clouds in \cref{fig:main_results}, which are color-coded according to the Chamfer distance.

\subsection{Baselines}

\textbf{Mesh-Based Baselines:}
3DMMs are the most common model prior for 3D face tracking. Therefore, we compare against DECA~\cite{DECA} and the MICA tracker~\cite{MICA_ECCV2022}.
The former is a CNN that is trained in a self-supervised fashion on in-the-wild images to predict FLAME~\cite{FLAME} parameters. %
The latter is a state-of-the-art face tracker, inspired by Face2Face~\cite{thies2016face}.
Additionally, we compare against Neural Head Avatar (NHA)~\cite{nha}, which learns person-specific face offsets and expression dependent neural textures.

\textbf{Field-Based Baselines:}
IMavatar~\cite{IMavatars} uses neural fields to explain details beyond an underlying FLAME model, which is used as guidance during optimization.
HeadNeRF~\cite{hong2021headnerf} is a NeRF-based ~\cite{mildenhall2021nerf} neural 3DMM. To achieve high-fidelity appearance it relies on a screenspace CNN.

\begin{table}[h]
    \centering
    \setlength{\tabcolsep}{3pt}
    \begin{tabular}{p{0.27\linewidth}ccc}
			\toprule
			Method&\multicolumn{1}{c}{\small{$L_1$-Chamfer $\downarrow$}}&\multicolumn{1}{c}{\small{N. C. $\uparrow$}}&\multicolumn{1}{c}{\small{Recall@2.5mm $\uparrow$}}\\
			\midrule
            \small{DECA}~\cite{DECA}&$0.0034$  &$0.917$&$0.644$\\
			\small{MICATracker}~\cite{MICA_ECCV2022}&$0.0030$  &$0.932$&$0.654$\\
            \small{NHA}~\cite{nha}&$0.0055$  &$0.872$&$0.490$\\

            \small{IMavatar}~\cite{IMavatars}&$0.0054$  &$0.888$&$0.625$\\
            \small{HeadNeRF}~\cite{hong2021headnerf}&$0.0049$  &$0.883$&$0.504$\\
			Ours& $\textbf{0.0024}$  &$\mathbf{0.940}$&$\mathbf{0.785}$\\
   \bottomrule
		\end{tabular}	
	\vspace{-0.2cm}	
    \caption{Quantitative comparison of 3D face reconstruction from RGB videos. The chamfer distance is reported in meters.}
    \label{tab:main_results}
    
    \vspace{-0.2cm}
    
\end{table}

\begin{figure}[htb]
    \vspace{-0.2cm}
    \centering
    \includegraphics[width=\columnwidth]{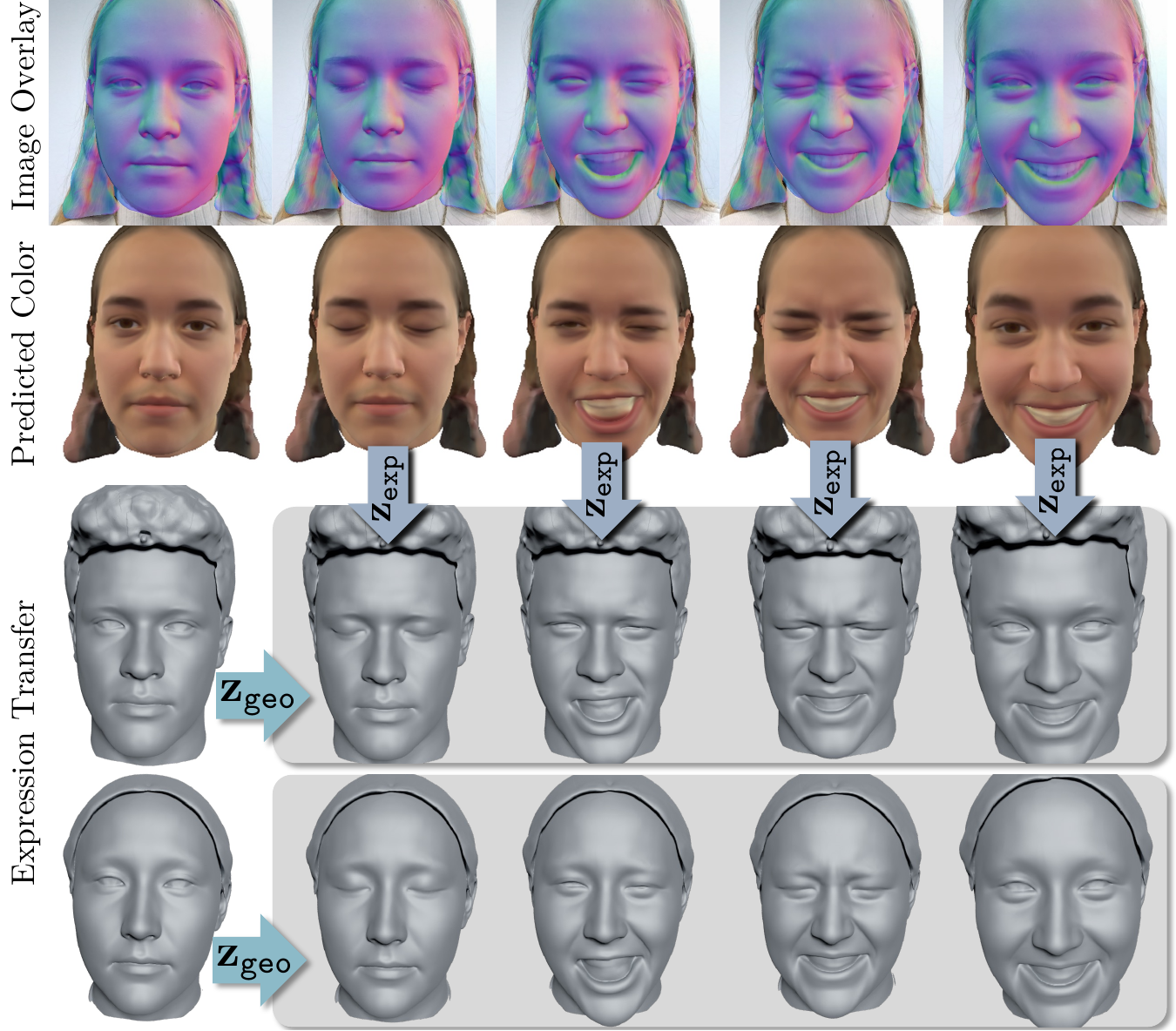}
    \vspace{-0.6cm}
    \caption{
    \textbf{Cross-Reenactment:} 
    We show five reconstructions from same sequence (top two rows). 
    A 50-50 overlay (top row) of reconstructed normals and the RGB images shows accurate image-space-alignment. 
    The bottom two rows show cross-reenactment by transferring expression codes to two other subjects.
    }
    \label{fig:overlay}
    \vspace{-0.2cm}
\end{figure}

\subsubsection{Implementation Details}

\textbf{Training \OURS}
We implement our model in Pytorch~\cite{pytorch} and utilize PytorchGeometric~\cite{pytorch_geometric} to restrict computations in canonical space to the $k$ nearest anchors. 
We follow NPHM~\cite{giebenhain2023nphm} and use 64 dimensions for the global parts of $\zapp$ and $\zexp$. For the local codes we use 32 dimensions. For the expression codes $\zexp$ we use 100 dimensions.
We train our model for 2500 epochs, use a batch size of 64, and a learning rate of $5e^{-4}$ for the networks and $2e^{-3}$ for the latent codes. We train on the updated release of the NPHM dataset ~\cite{giebenhain2023nphm} using 4 NVIDIA RTX2080 GPUs with 12GB of VRAM taking roughly 52 hours until convergence. More details are provided in our supplementary.

\textbf{Data Pre-Processing}
We perform several common pre-processing steps to remove parts of the observed images that are not included in our learned prior. %
Namely, we rely on face detection~\cite{deng2020retinaface}, facial landmark detection~\cite{PIPnet}, semantic segmentation to remove the torso \cite{zheng2022farl}, as well as, video matting \cite{ke2022modnet} to remove the background.
For all baselines, we follow their proposed pre-processing pipeline.

\textbf{Tracking}
For each step of SGD we randomly sample 500 rays. During volume rendering, we randomly sample 32 coarse samples, and additional 32 samples using importance sampling. We start with a large variance for the NeuS~\cite{wang2021neus} rendering, which is decayed over time to concentrate tightly around the surface. 
We perform 250 optimization steps for the first frame, and 60 steps per frame otherwise. To build forward correspondences for our landmark loss, we use 5 random initializations for iterative root finding. 
More details are provided in the supplementary.

Our optimization operates at roughly 1.2 frames per minute.
As a comparison, the MICA tracker can track 2 frames per minute using the default settings, and IMavatar operates at roughly 0.4 frames per minute.

\subsection{Tracking Results}
\label{sec:main_results}

We compare \OURS to our baselines by fitting each model to all the 20 monocular RGB sequences individually. 
\begin{figure}[htb]
    \vspace{-0.2cm}
    \centering
    \includegraphics[width=\columnwidth]{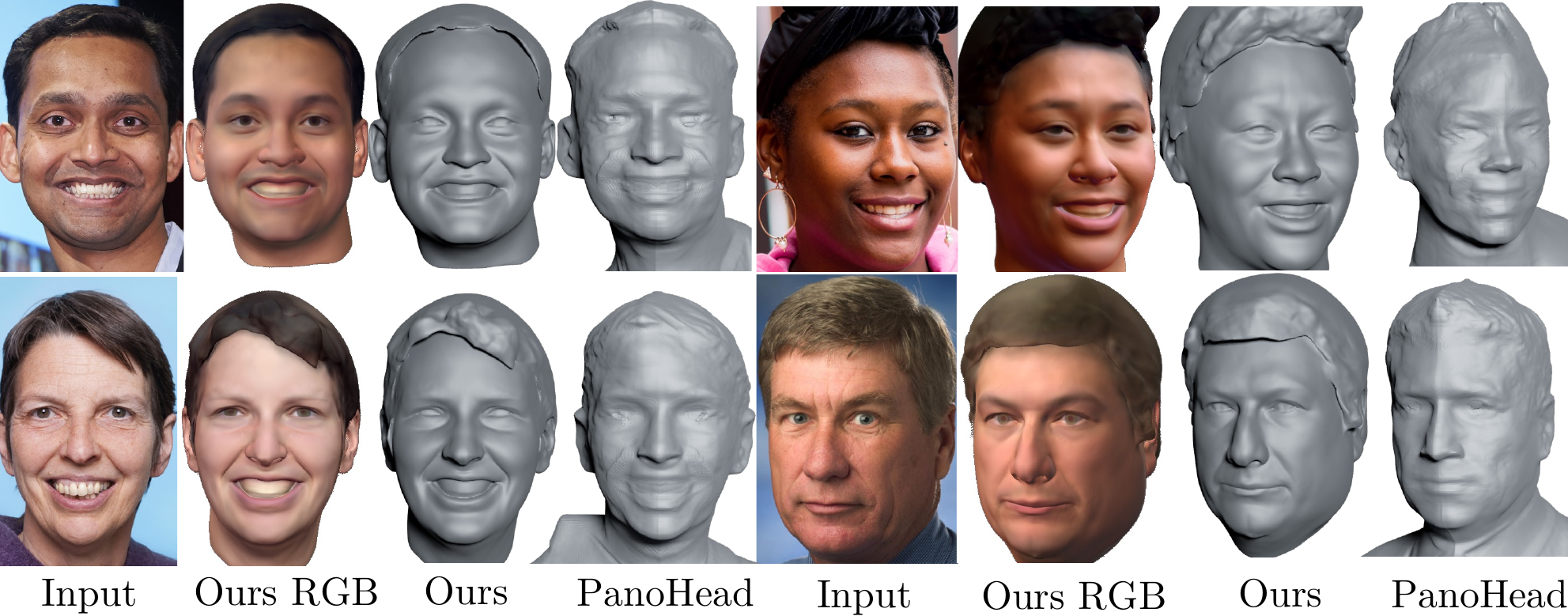}
    \vspace{-0.6cm}
    \caption{\textbf{Single-Image Reconstructions} on FFHQ~\cite{karras2019stylgan+ffhq} images.
    }
    \vspace{-0.4cm}
    \label{fig:ffhq}
\end{figure}
We report quantitative and qualitative results in \cref{tab:main_results} and \cref{fig:main_results}, respectively. For results on the complete sequences, we kindly refer to our supplementary video.
\cref{fig:main_results} shows that {\OURS} reconstructs important details about the face shape and expressions, that significantly help to recognize the identity and interpret the reconstructed emotion correctly.
Compared to the 3DMM-based approaches, DECA and the MICA tracker, {\OURS} is capable of reconstructing complete heads, including the mouth inside and hair.
IMavatar, on the other hand, suffers from its increased representational capacity compared to FLAME, due to the difficulty of task.
Both NHA and HeadNeRF employ high capacity neural networks for high-fidelity renderings. Consequently, the geometry of these approaches is under-constrained.
The quantitative evaluation reported in \cref{tab:main_results} confirms these findings. 

Additionally, we show qualitative results for five frames of the same sequence in \cref{fig:overlay}, to demonstrate temporal consistency and the alignment of our reconstructed geometry against the input sequence in screen space. 
Alongside, we show the predicted color images $\hat{I}$ of {\OURS}, to give further insights into our rendering loss $\mathcal{L}_{\texttt{RGB}}$, which mainly drives our optimization.
Finally, we perform cross-reenactment by transferring the reconstructed latent codes $\zexp$ to the identity codes $\zgeo$ and $\zapp$ from another participant.
Visually, the resulting reenactments capture the contents of the original expression to a high degree.

\textbf{Single-Image Reconstruction}
Furthermore, in \cref{fig:ffhq}, we demonstrate \OURS's 3D reconstruction capabilities on single in-the-wild images from the FFHQ dataset~\cite{karras2019stylgan+ffhq}.
This indicates that our learned prior is strong enough for sparse observations, extreme lighting conditions and diverse identities. 
Additionally, we include a qualitative comparison against PanoHead~\cite{An_2023_panohead}, a recent SOTA 3D generative head model trained on the FFHQ dataset.

\subsection{Ablations}
\label{sec:ablations}

 \begin{table}[H]
 \vspace{-0.2cm}	
 \centering
 \begin{tabular}{c|p{0.27\linewidth}ccc}
 \hline
\multicolumn{2}{p{0.27\linewidth}}{\quad ~~ Method} & \multicolumn{1}{c}{$L_1$-Ch. $\downarrow$} & \multicolumn{1}{c}{N. C. $\uparrow$} & \multicolumn{1}{c}{Recall $\uparrow$} \\
 \hline
 \parbox[t]{2mm}{\multirow{3}{*}{\rotatebox[origin=c]{90}{\footnotesize{tracking}}}} & \small{w/ sphere tracing} &$0.0033$  &$0.905$&$0.674$\\
 & \small{w/o spher. harm.} &$0.0028$  &$0.923$&$0.718$\\
 & \small{w/o $\mathcal{L}_{\texttt{lm}}$} &$0.0027$  &$0.939$&$0.745$\\
 \hline
 \parbox[t]{2mm}{\multirow{4}{*}{\rotatebox[origin=c]{90}{\footnotesize{architecture}}}} & 
 \small{NPHM$_{\text{app}}$} & $0.0028$  &$0.926$&$0.724$\\
 & \small{w/ $|\mathbf{a}|=39$} & $0.0027$  &$0.934$&$0.761$\\
 & \small{w/o color comm.} &$0.0028$  &$0.933$&$0.735$\\
 & \small{w/ global MLP} &$0.0026$  &$\mathbf{0.940}$&$0.768$\\
\hline
\hline
\multicolumn{2}{p{0.27\linewidth}}{\quad ~~ \textbf{Ours}}& $\textbf{0.0024}$  &$\mathbf{0.940}$&$\mathbf{0.785}$ \\
\hline
\end{tabular}
	\vspace{-0.2cm}	
    \caption{Ablations on single components of our tracking approach (first 3 rows), and of our architecture (second section). %
    }
    \label{tab:ablation_results}
    	\vspace{-0.2cm}	

 \end{table}

\newcolumntype{Y}{>{\centering\arraybackslash}X}
\newcolumntype{P}[1]{>{\centering\arraybackslash}p{#1}}
\begin{figure}[htb]
    \centering
    \includegraphics[width=\columnwidth]{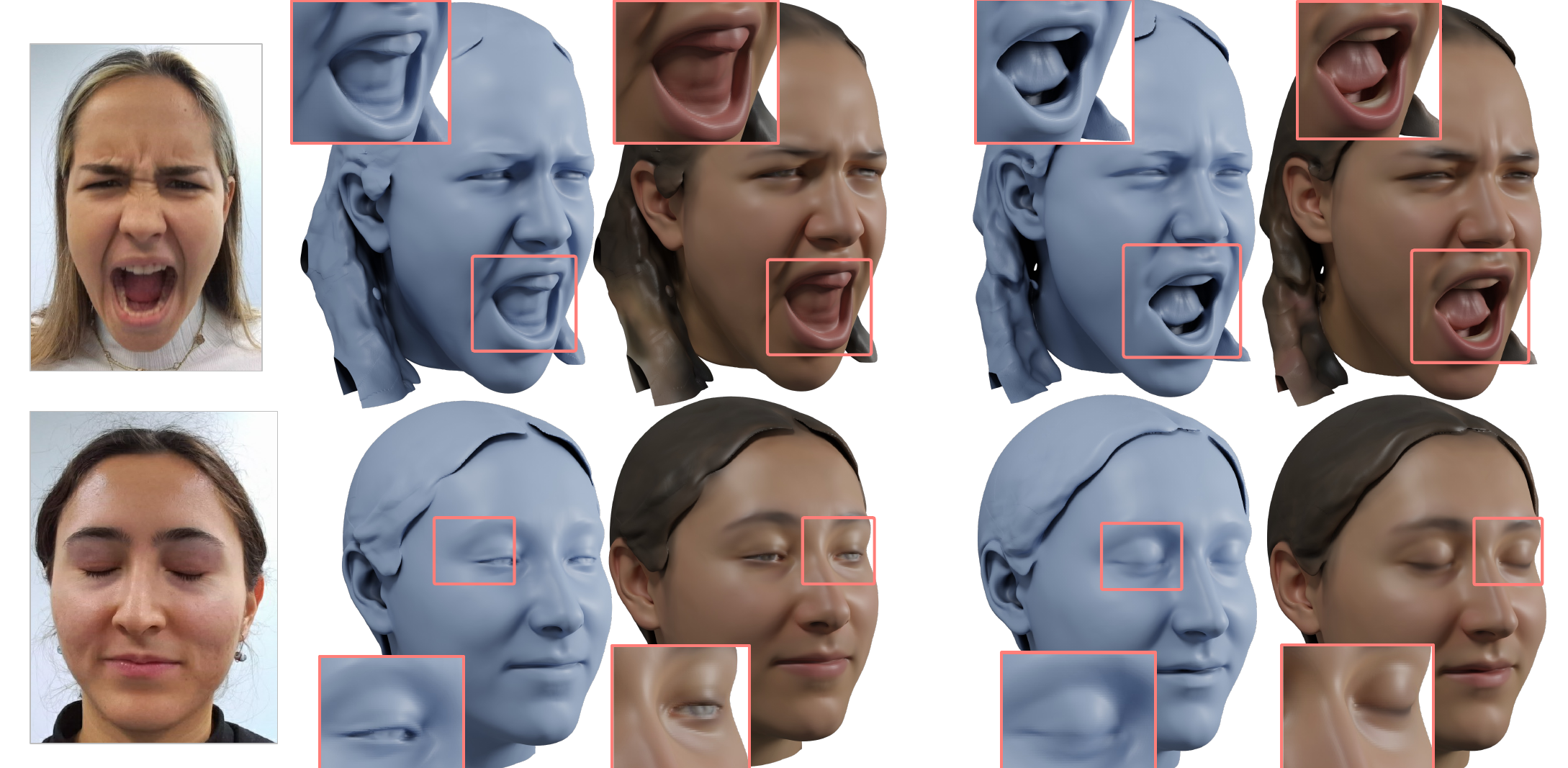}
     \begin{tabularx}{\columnwidth}{
         P{0.15\columnwidth}%
         P{0.33\columnwidth}%
         P{0.33\columnwidth}%
     }
    \small{RGB}   & \small{wo/ hyper-dims.} & \small{w/ hyper-dims.}\\
    \end{tabularx}
    \vspace{-0.6cm}
    \caption{\textbf{Effect of Hyper-Dimensions}: Without the addition of hyper-dimensions, the backward deformation model cannot accurately reconstruct geometry and color for topologically challenging expressions, e.g. the opening of mouth and closing of eyes.}
    \label{fig:effect_hyper}
    \vspace{-0.4cm}
\end{figure}

We support several of our claims by ablations on the same 20 Kinect sequences. Quantitative results are reported in \cref{tab:ablation_results} andqualitative results in our supplemental.

\textbf{Tracking Algorithm. ~} 
Firstly, we show that NeuS-style volume rendering \cite{wang2021neus}, instead of the sphere-tracing-based render for implicit surfaces of IDR~\cite{yariv2021idr}, is essential for the success of our tracking approach.
Second, the use of spherical harmonics is crucial to account for lighting conditions that vastly differ from our training data. Otherwise, the RGB loss is dominated by lighting effects that our model cannot explain.
Lastly, we note that without using the landmark loss $\mathcal{L}_{\texttt{lm}}$ our optimization often performs similarly well, but tends to get stuck in local minima, if large and fast mouth movements, e.g. during shouting, are encountered.

\textbf{Difference to NPHM~\cite{giebenhain2023nphm}. ~}
To analyze differences to the NPHM model, we train an extended version of NPHM, denoted as NPHM$_{\texttt{app}}$, which predicts color in the canonical space identical to {\OURS}, and uses our anchor layout. The main remaining differences are our use of backward (instead of forward) deformations and hyper dimensions. 
Additionally, we ablate the effect of our anchor layout by training {\OURS} using NPHM's anchor layout consisting of $|a|=39$ anchor points. This leads to a less effective landmarks loss and slightly reduced model capacity.

\textbf{The effect of $\mathbf{h}_{\texttt{geo}}$. ~}
Furthermore, we show the significance of removing the communication channel between geometry and color networks. To this end, we train a model that uses canonical coordinates $x_c$ instead of $\mathbf{h}_{\texttt{geo}}(x_c)$ as input to the local color MLPs defined in \cref{eq:color_communication}. We hypothesize that for such a model, gradients through our most important loss $\mathcal{L}_{\texttt{RGB}}$ are less informative for $\zgeo$ and $\zexp$.

\textbf{Local vs Global MLPs. ~}
Additionally, we ablate the effect of using the local MLP ensemble from NPHM~\cite{giebenhain2023nphm} against a simpler architecture, that represents $\Fgeo$ and $\Fapp$ using a global MLP. 
To account for good tracking of extreme expressions, 
we find that the landmark loss is equally important for this global architecture.
To this end, we include the anchor prediction MLP $\mathcal{A}$ into this ablation experiment, such that the usage of $\mathcal{L}_{\texttt{lm}}$ becomes viable. 
Doing so, we are able to associate $\mathbf{a}_c$ with $\zgeo$, and achieve good tracking performance, with slightly fewer geometric details, as reflected in the metrics.

\textbf{Effect of Hyper-Dimensions. ~}
Finally, in \cref{fig:effect_hyper} we highlight the importance of using hyper-dimensions for correctly representing topologically challenging expressions, such as the closing of the eyes and opening of the mouth. Note that the metrics are barely affected, since mouth and eye regions are excluded due to strong sensor noise.

\section{Limitations}

In our experiments, we show that \OURS~can reconstruct high-quality human heads from monocular videos; however, at the same time, we believe that there are still several limitations and opportunities for future work.
For instance, while spherical harmonics can be used to account for simple lighting conditions without increasing the model complexity, we believe that reconstructions could be improved by addressing lighting and shadows more thoroughly. 
Possible options are the inclusion of a more advanced shading model during volume rendering ~\cite{neural-3d-relightable_facelit}, image-space delighting \cite{yeh2022learning_lumos}, as well as, CNN-based image encoders \cite{DECA, continuous_landmarks}.
Another limitation is our tracking speed. While this is partially explained by our unoptimized implementation that runs a full optimization for each frame, we believe that several advances can be made, e.g. using CNN-based initialization \cite{pavllo2023shape}, coarse-to-fine optimization, faster neural-field backbones \cite{mueller2022instant, Chen2022ECCV_tensorf} and second-order optimization for tracking~\cite{thies2016face}.

\section{Conclusion}

In this work we have introduced \OURS, a neural-field-based parametric face model, that represents faces using an SDF and texture field in canonical space, and represents movements using backward deformations, augmented with hyper-dimensions. We enforce a tight communication between appearance and geometry to facilitate efficient inverse rendering.
By including explicit control points in our implicit geometry representation, we have developed a highly accurate 3D face tracking algorithm based on volumetric rendering for implicit surfaces.
\OURS~achieves significantly more accurate 3D reconstruction on challenging monocular RGB videos, compared to all our baselines.
We believe that our work makes the use of neural parametric head models much more accessible for many downstream tasks. 
We hope that our work inspires more research to explore the use of neural-field-based parametric models and develop the necessary toolsets that are already available for classical 3DMMs.

\subsubsection*{Acknowledgements}
This work was funded by Synthesia and supported by the ERC Starting Grant Scan2CAD (804724), the German Research Foundation (DFG) Research Unit ``Learning and Simulation in Visual Computing''. We would like to thank our research assistants Mohak Mansharamani and Kevin Qu, %
and Angela Dai for the video voice-over.

\newpage 

{
    \small
    \bibliographystyle{ieeenat_fullname}
    \bibliography{main}
}

\newpage

\appendix

\huge \textbf{Appendix}
\normalsize

\section{Overview}

This supplementary document provides additional implementation details on our network architecture (\cref{sec:network_arch}), training (\cref{sec:training}) and  tracking strategy (\cref{sec:tracking}).

Additionally, we present more qualitative results (\cref{sec:additional_results}) and discuss our ablation experiments (\cref{sec:ablations}).

We kindly suggest the reviewers watch our supplementary video, for a temporally complete visualization of the tracked sequences.

\section{Implementation Details}
\label{sec:add_impl_details}

In \cref{sec:network_arch} we provide details about the individual network components of \OURS. \cref{sec:efficient_impl} describes how we implement a memory efficient variant of the MLP ensemble proposed in \cite{giebenhain2023nphm}.

\subsection{Network Architectures}
\label{sec:network_arch}

Some of mentioned details in this subsection require detailed knowledge about NPHM~\cite{giebenhain2023nphm}.

\paragraph{Expression Network}
To represent our backward deformation field $\Fexp$ we use a 6-layer MLP with a width of 400. The expression codes $\zexp$ are 100 dimensional. The dependence on $\zgeo$ is bottlenecked by a linear projection to 16 dimensions, as proposed in \cite{giebenhain2023nphm}.

\paragraph{Geometry Network}
Our local geometry MLPs $f_{\texttt{geo}}^k$ have 4 layers and a width of 200. Out of the 65 anchors, 30 are symmetric, meaning that the ensemble consists of $64 - 30 = 34$ MLPs.
Note, however, that the spatial input of $f_{\texttt{geo}}^k$ is augmented with the predicted hyper-dimensions.

\paragraph{Appearance Network}
Our appearance MLPs $f_{\texttt{app}}^k$ follow the same structure as $f_{\texttt{geo}}^k$, but receive extracted geometry features $\mathbf{h}_{\texttt{geo}}(x_c)$ as input. $\mathbf{h}_{\texttt{geo}}$ is a two-layer MLP (widths 100 and 16), that maps the hidden features of the last layers of $f_{\texttt{geo}}^k$ to 16 dimensions.

\paragraph{Anchor Prediction}
Compared to the anchor layout used in NPHM~\cite{giebenhain2023nphm}, we increase the number of anchors from 39 to 65, and rearrange them, such that the anchors coincide with the most important facial landmarks for tracking. \cref{fig:rescale} shows our anchor layout.
The anchor prediction MLP $\mathcal{A}$ consists of 3 linear layers and has a hidden dimension of 64.

\subsection{Efficient Implementation}
\label{sec:efficient_impl}
To account for the computational burden of the increased number of anchors and added appearance MLPs, we prune the computations of the local MLP ensemble. 

\paragraph{$k$NN Pruning}
NPHM executes every MLP $f_{\texttt{geo}}^k$ for each query point $x_c$.
Instead, we use Pytorch3D~\cite{ravi2020pytorch3d} to compute the 8 nearest neighbors $\mathcal{N}_{x_c}$ for each query.
Then, we conceptualize the execution of local MLPs as a graph convolution, implement usingPytorchGeometric~\cite{pytorch_geometric}. The graph convolution is restricted to $\mathcal{N}_{x_c}$ (see equation 2 in the main document).
In practice, this decreases the number of MLP executions for each query from 65 to 8 (the number of nearest neighbors). Hence,  GPU memory demand is roughly reduced 8-fold

\begin{figure}[htb]
    \centering
    \includegraphics[width=\columnwidth]{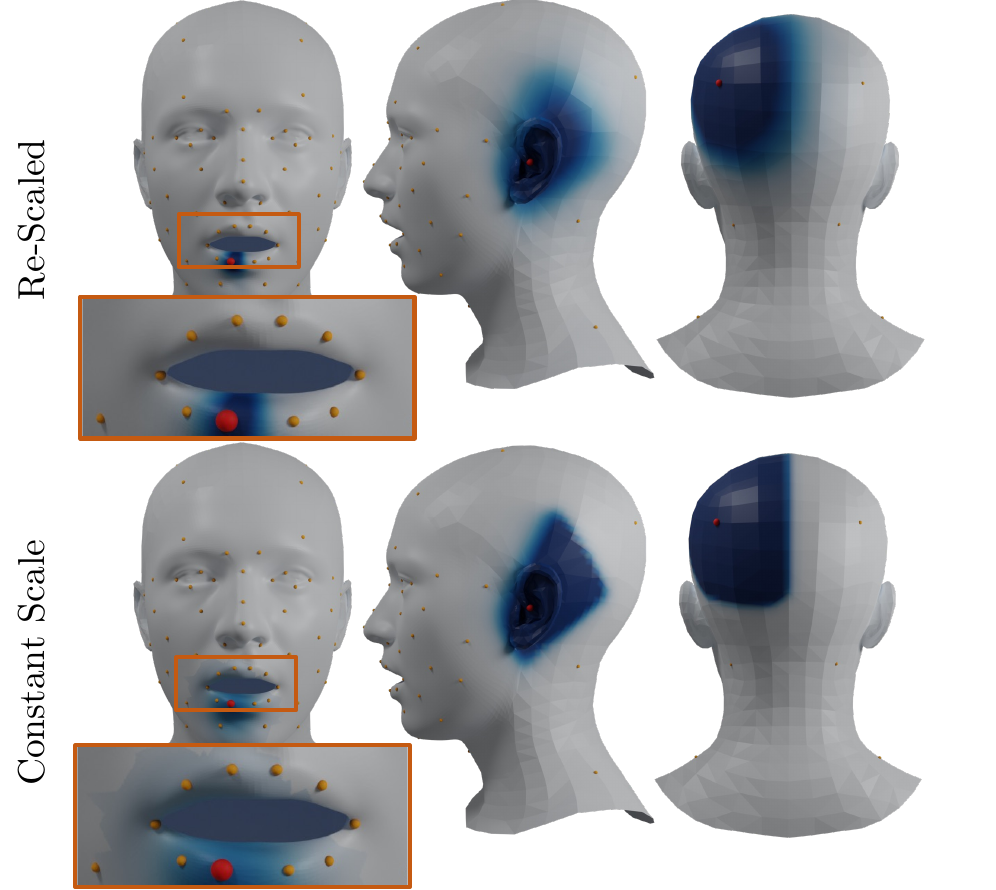}
    \caption{
    \textbf{Re-Scaling $w_k$:} We show weights $w_k$ for three different anchors (red) among all the 65 anchors (orange). The mesh surface are colored according to $w_k$ where white corresponds to a low value and blue to a large value. The top row shows our re-scaled weights compared to a constant scale (bottom row). Note the discontinuities on the bottom left, and the sharp decay on the bottom right.
    }
    \label{fig:rescale}
\end{figure}

\paragraph{Re-Scaling $w_k$}
For a given query point $x_c$ and local MLP associated to the anchor point $\mathbf{a}_k$, NPHM uses weights

\begin{equation}
    w_k^*(x_c, \mathbf{a}_k) = \text{exp}\left( \frac{-\Vert x_c-\mathbf{a}_k\Vert_2}{ 2\sigma }\right)
    \label{eq:const_weight},
\end{equation}

 and normalizes them to $w_k$ in order to blend the predictions of the individual MLPs.
However, when restricting the computations to the set of nearest neighbors, such a constant-scale Gaussian weighting results in discontinuous for points on the boundary of Voronoi cells, i.e. when the set of nearest neighbors changes. 

As demonstrated in the bottom left of \cref{fig:rescale}, the influence of the highlighted anchor point exhibits a sharp boundary. 
This effect can be mitigated by reducing $\sigma$ to be significantly smaller than the size of the Voronoi cells. However, due to the non-uniform spatial arrangement of anchors, finding a single $\sigma$ that ensures smooth boundaries for all anchors is impossible.

Consequently, we vary 
\begin{equation}
    \sigma(x_c) = \frac{1}{4} \underset{x\in \mathcal{N}_{x_c}}{\mathrm{max}} \Vert x_c - x\Vert_2
\end{equation}
according to the set of nearest neighbors of $\mathcal{N}_{x_c}$.
Doing so ensures that $w_k^*(x_c, \mathbf{a}_k)$ decays quickly enough to zero when approaching the boundaries of its Voronoi cell.

\subsection{Training Details}
\label{sec:training}

\subsubsection{Data Preparation}

We use the 3D textured scans of the NPHM dataset \cite{giebenhain2023nphm} for training.
To this and we sample points on the surface $\mathcal{S}_{\text{surf}}$ and near the surface $\mathcal{S}_{\text{near}}$, and define $\mathcal{S}_{\text{all}} = \mathcal{S}_{\text{surf}} \cup \mathcal{S}_{\text{near}}$.
For $x_p \in \mathcal{S}_{\text{all}}$ we precompute its normal $n(x_p)$ and color $\text{RGB}(x_p)$.
Additionally, we precompute samples $(x_p, x_c) \in \mathcal{S}_{\text{corr}}$ of corresponding points in posed and canonical space following \cite{NPM} and using the provided registered meshes in the NPHM dataset.

\subsubsection{Loss Functions}

 We train {\OURS} in an end-to-end fashion, similar to ImFace~\cite{ImFace} which jointly trains geometry and expression networks.

\paragraph{Geometry Supervision}

The employed losses are similar to \cite{IGR}, however, adopted to dynamic objects similarly to \cite{ImFace}. Hence, the main losses for the geometry and expression supervision put constraints on the zero-level set through
\begin{equation}
    \mathcal{L}_{\text{level-set}} = \sum_{x_p \in \mathcal{S}_{\text{surf}}} \Vert \Fgeo(\Fexp(x_p)) \Vert_1
\end{equation}
and on the surface normals through
\begin{equation}
    \mathcal{L}_{\text{n}} = \sum_{x_p \in \mathcal{S}_{\text{surf}}} \Vert \nabla_{x_p}\Fgeo(\Fexp(x_p)) - n(x_p) \Vert_2,
\end{equation}
where we omit the dependence on latent codes for brevity. Additionally, we enforce the eikonal constraint
\begin{equation}
    \mathcal{L}_{\text{eik}} = \sum_{x_p \in \mathcal{S}_{\text{all}}} \Vert \nabla_{x_p}\Fgeo(\Fexp(x_p)) - 1 \Vert_2.
\end{equation}
To guide $\Fexp$ during the first half of training we include a correspondence loss
\begin{equation}
    \mathcal{L}_{\text{corr}} = \sum_{(x_p, x_c) \in \mathcal{S}_{\text{corr}}} \Vert \Fexp(x_p) - x_c \Vert_1.
\end{equation}
On one side this provides direct expression supervision. On the other side $\mathcal{L}_{\text{corr}}$ also enforces the first 3 dimensions of the canonical space to behave as Euclidean as possible. This is not only desirable but also extremely important for the landmark loss $\mathcal{L}_{\texttt{lm}}$ to work.
For the same reason, we regularize predicted hyper-dimensions $\omega = [{\Fexp}(x_p)]_{\omega}$ to be small using 
\begin{equation}
    \mathcal{L}_{\text{hyper}} = \sum_{x_p\in \mathcal{S}_{\text{all}}} \Vert ~ [{\Fexp}_{\omega}(x_p) ]_{\omega} ~ \Vert_2.
\end{equation}
In a similar fashion, we regularize predicted deformations to be small
\begin{equation}
    \mathcal{L}_{\text{def}} = \sum_{x_p \in \mathcal{S}_{\text{surf}}} \Vert \Fexp(x_p) - x_p \Vert_2.
\end{equation}
Finally, we include the same regularization terms as \cite{giebenhain2023nphm}, i.e. we constrain the norm of $\zgeo$ and $\zexp$ and apply a symmetry loss on the symmetric parts of $\zgeo$.

\paragraph{Anchor Supervision}
Anchor positions are directly supervised using
\begin{equation}
    \mathcal{L}_{\mathcal{A}} = \Vert \mathbf{a}_{\text{gt}} - \mathcal{A}(\zgeo)\Vert_F
\end{equation}
where the ground truth anchor positions $\mathbf{a}_{\text{gt}}$ are extracted from the registered meshes in FLAME~\cite{FLAME} topology, as provided by the NPHM dataset. 
Therefore, the anchors are supervised to follow the Euclidean coordinate system of the FLAME model. While this seems obvious, we note that without the necessary precautions imposed by $\mathcal{L}_{\text{corr}}$, $\mathcal{L}_{\text{def}}$, and $\mathcal{L}_{\text{hyper}}$, our canonical space becomes non-euclidean, similarly to \cite{neural_volumes, Lei2022CaDeX, Zheng2023pointavatar}.

\paragraph{Appearance Supervision}

The appearance codes $\zapp$ and network $\Fapp$ are jointly optimized alongside the geometry, by including
\begin{equation}
    \mathcal{L}_{\text{app}} = \sum_{x_p \in \mathcal{S}_{\text{all}}} \Vert \Fapp(\mathbf{h}_{\texttt{geo}}(\Fexp(x_p) ) - \text{RGB}(x_p) \Vert_1
\end{equation}
into our training. Similarly as before, we also regularize the norm of $\zapp$.
We do not include a perceptual loss during training, as done in \cite{phomoh, lin2023ssif}, since we are focused on geometry reconstruction via inverse rendering, instead of photorealistic appearance.

\subsubsection{Training Strategy}
Using the above-mentioned losses, we train all networks and latent codes jointly in an auto-decoder fashion \cite{park2019deepsdf}. We use the Adam optimizer \cite{adam}, and periodically divide the learning rates by half every 500 epochs, for a total of 2500 epochs and use a batch size of 64. We start with $lr_{\text{networks}}=0.0005$, $lr_{\text{lat-can}} = 0.002$ and $lr_{\text{lat-exp}} = 0.01$, for the network parameters, latent codes for canonical space and latent expression codes, respectively.

\subsection{Tracking Details}
\label{sec:tracking}

We perform iterative root finding using 5 random samples normally distributed around the canonical anchor $\mathbf{a}_k$ of interest, as we experience similar convergence issues to \cite{chen2021snarf} that are dependent on the initial position.

Since the inside of the mouth is subject to extreme shadows, far beyond what our simple lighting assumptions can explain, we use the predicted facial segmentation masks \cite{FaRL} to down-weigh the color loss $\mathcal{L}_{\texttt{RGB}}$ by a factor of 25 for that region.

Furthermore, we employ several mechanisms to encourage a \emph{coarse-to-fine} optimization. 
First, we decay all learning rates of the employed Adam optimizer periodically throughout the optimization. The learning rate for the head pose and spherical harmonics parameters $\zeta$ start larger and decay faster compared to the learning rate of the latent codes.
Second, we increase the inverse standard deviation from the NeuS~\cite{wang2021neus} volume rendering formulation from $0.3$ to $0.8$. Therefore, the rendering densities are initially distributed widely around the surface, allowing for a large volume that receives gradients in the coarser stages of optimization. 
Third, the influence of the landmark loss $\mathcal{L}_{\texttt{lm}}$ is strongly decayed throughout the optimization progress. Initial epochs strongly rely on landmark guidance, while later ones are barely affected by it anymore. Additionally, we weigh the landmarks of the eyes, mouth and chin $100$ more then the remaining ones.

\section{Additional Qualitative Results}
\label{sec:additional_results}

\subsection{Additional Comparisons}

\begin{figure*}[htb]
    \centering
    \includegraphics[width=\textwidth]{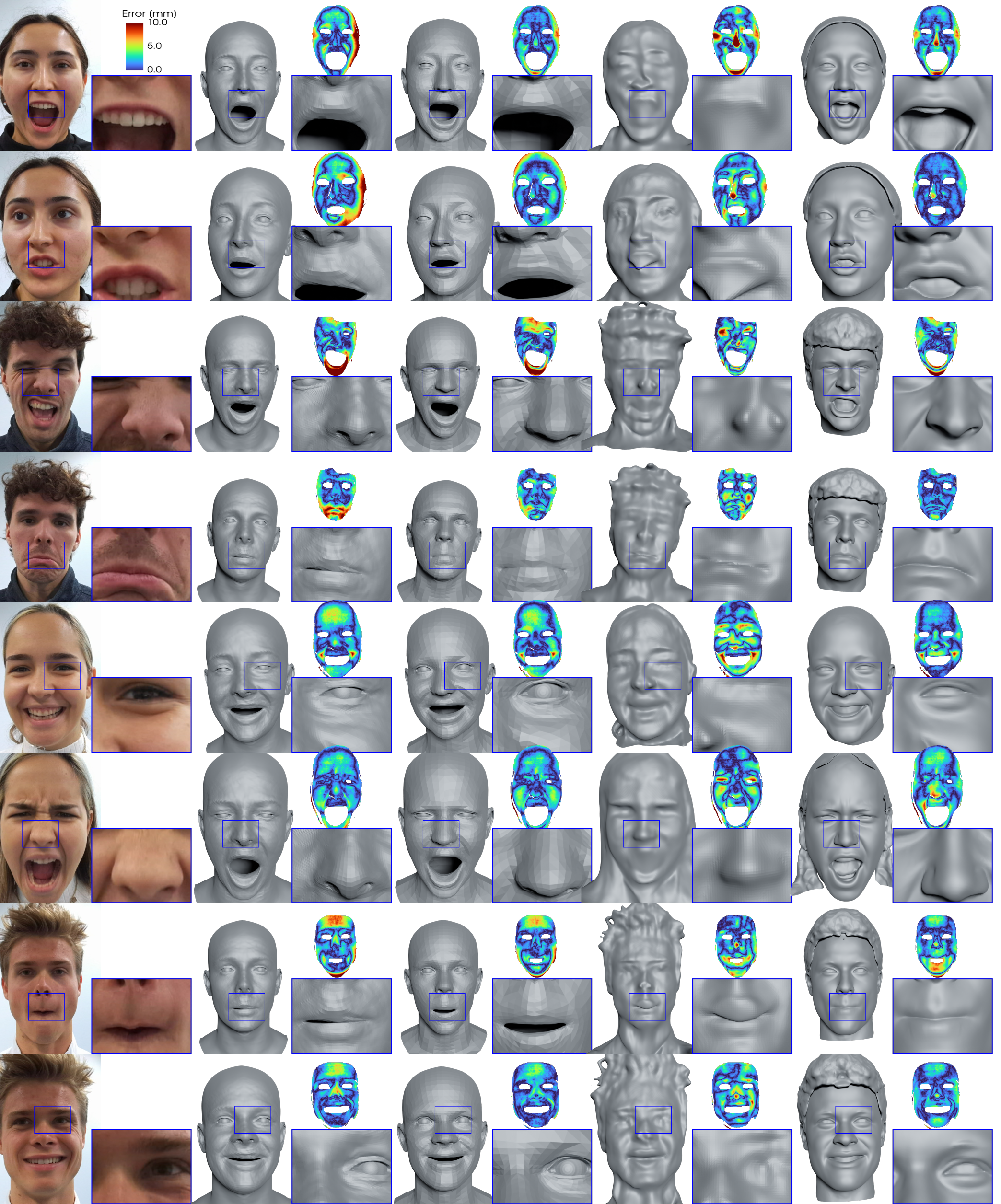}
    \caption{
    \textbf{Tracking Comparison:} We show additional qualitative results of the monocular 3D reconstruction task. The error maps show the color-coded point-to-mesh distance from the back-projected Kinect depth to the reconstruction.
    }
    \label{fig:additional_results}
\end{figure*}

Next to the results in the main paper and our supplementary video, we show additional qualitative comparisons against our baselines in \cref{fig:additional_results}. Note that each row shows a frame from a different sequence, which are reconstructed separately.

Note that we due not show additional results for NHA~\cite{nha} and HeadNeRF~\cite{hong2021headnerf}, since both methods do not have accurate geometry as their main focus.

\subsection{Ablations}
\label{sec:ablations}
\begin{figure*}[htb]
    \centering
    \includegraphics[width=\textwidth]{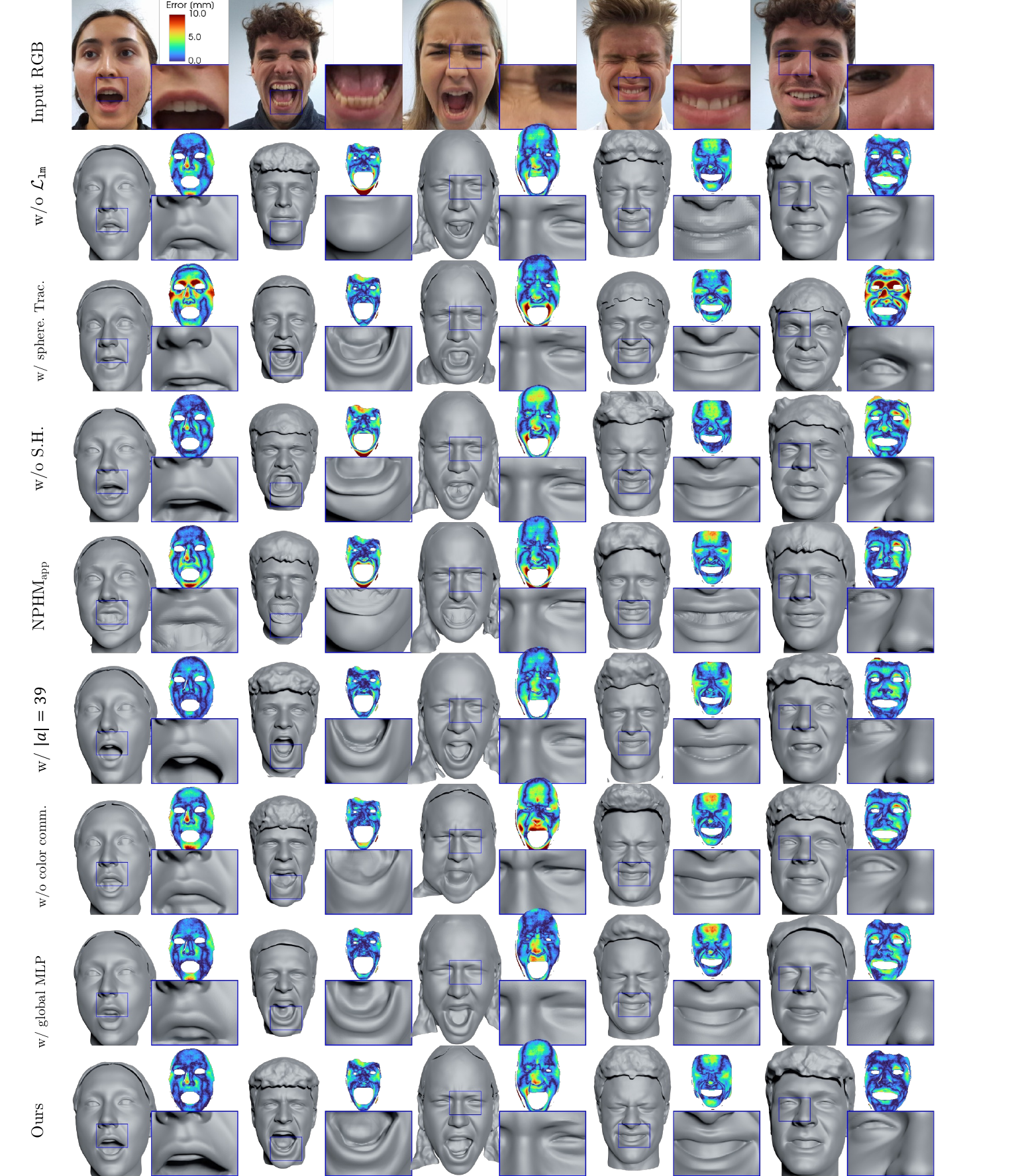}
    \caption{
    \textbf{Ablation Results:} Qualitative comparison of our ablation experiments, as quantitatively reported in Table 2 in the main document. Rows and columns are transposed compared to our other result figures. The error maps show the color-coded point-to-mesh distance from the back-projected Kinect depth to the reconstruction. See \cref{sec:ablations} for a description of our findings.
    }
    \label{fig:abl_results}
\end{figure*}

While our main document only reported quantitative results of our ablation experiments, due to space reasons, \cref{fig:abl_results} and our supplementary video show qualitative results. In the following we highlight some key insights from our ablation experiments:

\paragraph{Effect of $\mathcal{L}_{\texttt{lm}}$}
Generally, our tracking performs well even when the landmark loss is disabled. However, some extreme expressions are completely missed without it, see the second column in \cref{fig:abl_results}.\\
Additionally, utilizing a landmark detector trained on large image collections of in-the-wild images provides some robustness against lighting and shadow effects.

\paragraph{Volume Rendering vs. Sphere Tracing}
Utilizing sphere tracing \cite{yariv2021idr}, instead of a volumetric formulation \cite{wang2021neus}, for differentiable SDF-based rendering results in reconstructions that are perceptively dissimilar to the subject. Additionally, we note that the sphere tracing sometimes gets stuck in local minima, where it is not able to remove hair geometry in front of the forehead, see columns four and five.

\paragraph{Spherical Harmonics}
Since our model is trained on 3D scans, with albedo-like texture, accounting for lighting effects is important. Removing the spherical harmonics term, makes the task slightly ill-posed and generally results in worse reconstruction quality.

\paragraph{Deformation Formulation}
We ablate our deformation module, consisting of backward deformations and hyper dimensions,  against the forward deformation utilized in NPHM~\cite{giebenhain2023nphm}. To this end we extend NPHM's canonical space using our proposed approach to include color prediction. We denote this model as NPHM$_{\texttt{app}}$. Due to its invert deformation direction iterative-root-finding is required during rendering and not for the landmark loss. Another difference is that it needs to be trained in two stages according to \cite{giebenhain2023nphm}. Otherwise, the same losses and hyperparamters are used for tracking. 

\cref{fig:abl_results} indicates that the forward deformation module mainly has problems in the mouth region, e.g. with folded lips.

\paragraph{Anchor Layout}
Additionally, we ablate the proposed anchor layout against the version used in NPHM, which uses 39 anchors instead of our proposed 65 anchors. This mainly results in a slightly less dense landmark loss, and slightly reduced capacity, due to a lower number of local MLPs.

\paragraph{Color Communication}
Conditioning the color MLP $\Fapp$ directly on canonical spatial coordinates $x_c$ instead of geometry features $\mathbf{h}_{\texttt{geo}}(x_c)$, gives the model extra freedom since both outputs are less correlated. For example in column 3 this results in a failure to separate the hair and cheek. Additionally, such a communication bottleneck was found to be beneficial for disentangling the geometry and appearance latent spaces \cite{phomoh}.

\paragraph{Local vs. Global MLPs}

Our MLPs modeling the SDF and texture field follow the local structure proposed in \cite{giebenhain2023nphm}, i.e. we use an ensemble of local MLPs, each centered around its specific facial anchor points. Additionally, symmetric face regions are represented using the same MLP, but with mirrored coordinates. Our main motivation for choosing such an architecture are the facial anchors, which we exploit to formulate our landmark loss. 
We realized that it is also possible to use the same landmark loss while using global MLPs for both SDF and texture field. To this end, it is necessary to add the anchor prediction network $\mathcal{A}$ to the architecture, although the predicted anchors are not used anywhere else in that architecture. We find that training such a model is still capable of successfully associating the geometry code $\zgeo$ with plausible facial anchors. 
Nevertheless, the local MLP ensemble still learns a more detailed latent representation, which, for example, shows in the slightly blurry eye reconstructions in columns three and five.

\newpage

\end{document}